\documentclass[11pt]{article}

\usepackage[preprint]{acl}

\usepackage{times}
\usepackage{latexsym}

\usepackage{stfloats}

\usepackage{booktabs}
\usepackage{multirow}
\usepackage[table]{xcolor}
\usepackage{graphicx}
\usepackage[normalem]{ulem}

\usepackage[T1]{fontenc}
\usepackage[utf8]{inputenc}

\usepackage{microtype}

\usepackage{inconsolata}

\usepackage{graphicx}
\usepackage{booktabs}
\usepackage{multirow}
\usepackage{graphicx}
\usepackage[table]{xcolor}
\usepackage{tabularx}
\usepackage{amsmath}
\usepackage{amssymb}
\usepackage{bm}

\newcommand{\R}{\mathbb{R}}

\title{Rethinking Groups in Critic-Free RLVR}

\author{
  Yihong Wu\textsuperscript{1}\thanks{\ Equal contribution. Emails: \texttt{yihong.wu@umontreal.ca}, \texttt{liheng.ma@mail.mcgill.ca}} \quad
  Liheng Ma\textsuperscript{2,3}\footnotemark[1] \quad
  Lingfeng Xiao\textsuperscript{4} \quad
  Muzhi Li\textsuperscript{5} \\
  \textbf{Xinyu Wang}\textsuperscript{2} \quad
  \textbf{Yingxue Zhang}\textsuperscript{6} \quad
  \textbf{Jian-Yun Nie}\textsuperscript{1} \\
  \vspace{0.2em} \\
  \textsuperscript{1}Université de Montréal \quad
  \textsuperscript{2}McGill University  \quad
  \textsuperscript{3}Mila - Quebec AI Institute \quad \\
  \textsuperscript{4}University of Waterloo
  \textsuperscript{5}The Chinese University of Hong Kong \quad
  \textsuperscript{6}Huawei Noah's Ark Lab
}

\begin{document}
\maketitle
\begin{abstract}
Reinforcement learning (RL) has become a central paradigm for post-training large language models. Existing critic-free RL methods typically generate a group of rollouts for the same question to estimate value baselines for advantage computation. However, this design suffers from data inefficiency, group synchronization barriers, and inflexibility with structured rollouts. In this work, we revisit the role of the ``group'' and show that its underlying function is not merely to estimate baselines but to prevent false penalties on negative samples. Building on this insight, we propose \textit{negative token filtering}, a simple and effective strategy that enables stable single-rollout training. We apply it to two batch-level advantage methods, achieving comparable performance on reasoning tasks and stronger performance on agentic tasks relative to group-based RL techniques.
\end{abstract}

\newcommand{\rfpp}{$\text{RF++}$}
\newcommand{\rfppm}{$\text{RF++}_{w/\ \text{Baseline}}$}
\newcommand{\gs}[1]{[$#1$]}

\section{Introduction}
% background
Reinforcement learning (RL) has become the \textit{de facto} paradigm for post-training Large Language Models (LLMs) to enhance their capabilities. To avoid the computational and memory overhead of a separate critic network, as in critic-based methods like PPO~\cite{schulman2017proximal}, critic-free methods have been widely adopted. Most of these—such as GRPO~\cite{shao2024deepseekmath}, RLOO~\cite{ahmadian2024back}, and ReMax~\cite{li2023remax}—generate multiple rollouts per prompt and use the resulting group to estimate value baselines for advantage computation.

\begin{figure}[t!]
    \centering
\includegraphics[width=\linewidth]{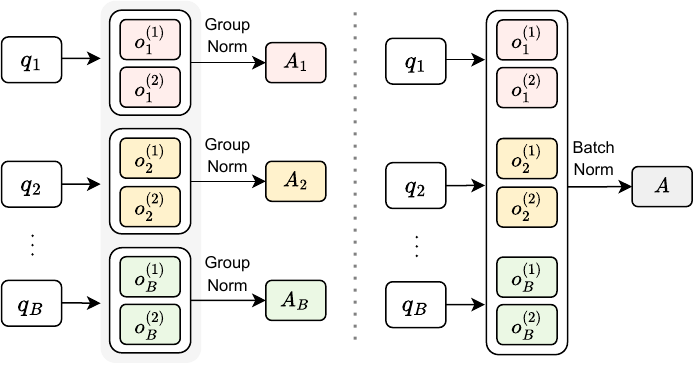}
    \caption{Advantage computation in GRPO vs. REINFORCE++. GRPO (left) computes advantages at the group level, whereas REINFORCE++ (right) computes them at the generation-batch level.}
    \label{fig:grpo_vs_reinforcepp}
\vspace{-1.5em}
\end{figure}
Although critic-free methods are more efficient than critic-based alternatives, their reliance on generating and grouping multiple rollouts still incurs a cost. For example, rollout grouping can introduce synchronization barriers~\cite{xu2025single}, force discarding groups whose rollouts receive identical rewards~\cite{yu2026dapo}, and prove inflexible for structured rollouts in agentic RL~\cite{feng2026group}. Therefore, recent works have explored critic-free methods that avoid multiple rollouts per prompt and group-based advantage computation~\cite{hu2025reinforce++, xu2025single}. REINFORCE++~\cite{hu2025reinforce++} replaces group normalization with batch-level normalization, computing advantages from generation batch rather than group-level statistics (Fig.~\ref{fig:grpo_vs_reinforcepp}).
However, for reasoning tasks it still typically uses a group size larger than one to improve training stability; with a single rollout, it suffers from severe training instability (Fig.~\ref{fig:neg_ablation}).
To fully enable single rollout generation, SPO~\cite{xu2025single} builds on batch-level normalization and introduce an additional tracker that estimates the value baseline via historical information.
However, this tracker requires extra sampling before training, adding computational overhead.

To uncover the role of grouping, we reverse-engineer the functional mechanism of rollout groups and use the resulting insight to develop an alternative single-rollout, critic-free policy optimization method. We make two observations: (1) the training instability of single-rollout RL originates from negative samples; and (2) using exactly one positive and one negative rollout per prompt restores stable training. We interpret these observations as follows.
First, an incorrect reasoning trajectory is rarely entirely wrong—it still contains many useful token patterns, such as formatting, intermediate reasoning steps, and tool-use cues.
Penalizing all of them equally therefore leads to harmful updates. Second, this harm is strongly alleviated when a positive trajectory for the same prompt is present: since the positive and negative rollouts typically share these functional tokens, the negative gradient on the shared tokens is partially cancelled out. In other words, the group does not merely estimate a baseline; it \textit{de facto} protects shared useful tokens from being over-penalized. We confirm this cancellation effect through (1) a statistical analysis of token overlap and (2) gradient projection onto the Top-$K$ subspace of the weight matrices.
Stemming from this insight, we propose a simple filtering strategy for negative trajectories that retains only the Top-$10\%$ lowest-probability tokens in the negative loss. We empirically verify this filtering on two batch-level advantage computation methods, and the resulting group-free methods outperform their group-based counterparts.

\begin{figure}
    \centering
    \includegraphics[width=.95\linewidth]{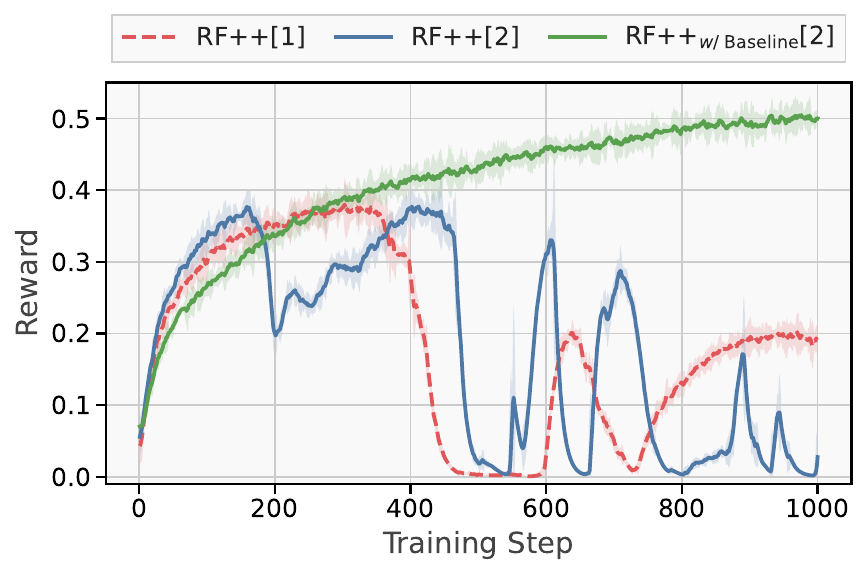}
    \vspace{-.5em}
    \caption{Training Curves of \rfpp\gs{1},~\rfpp\gs{2}~and \rfppm\gs{2}. This indicates that multi-rollout generation cannot guarantee stable training.
    The grouping mechanism is more critical for training stability,. 
    }
    \vspace{-1.5em}
    \label{fig:group_size_2}
\end{figure}
\section{Analysis}
\label{sec:understanding}

% In this section, we provide an in-depth analysis on the effects of multi-rollout sampling and group-based advantage estimation.

Most critic-free RL methods for LLMs introduce two mechanisms: multi-rollout sampling per prompt, and group-based advantage computation.
Note that a method with multi-rollout sampling might not introduce group-based advantage. 
On the other hand, the group-based advantage computation relies on multi-rollout sampling.
In this section, we disentangle these two mechanisms and
study their effects in isolation.

\subsection{The Effect of Grouping}
\label{sec:grouping}

% We first isolate group-based advantage by holding the number of rollouts fixed and varying only
% whether---and over how large a group---advantages are normalized.

We use REINFORCE++ (RF++)~\cite{hu2025reinforce++} as the base methods, which enables us to isolate the effects of multi-rollout generation and grouping.
Motivated by recent work~\cite{wu2025takes}, we focus on a group size $G$ of $2$ as the minimal multi-rollout setting, which allows a clean analysis while avoiding unnecessary experimental complexity.

Specifically, we compare three variants: \rfpp\gs{1},~\rfpp\gs{2}~and \rfppm\gs{2}, where \gs{n} denotes the group size of $n$.
The \rfpp~performs reward normalization over the mini-batch of rollouts, 
rather than within each prompt-level group as in GRPO.
The \rfppm~further subtracts the group-wise sample mean before batch-level normalization, thereby incorporating a GRPO-style grouping mechanism prior to mini-batch normalization -- a group of rollouts, which are all correct or all incorrect, will yield zero advantages and thus contributes no gradient to the optimization.

As shown in Fig.~\ref{fig:group_size_2}, although \rfpp\gs{2} is more stable than \rfpp\gs{1}, it still eventually collapses.
By contrast, \rfppm\gs{2} remains stable throughout training, with steadily increasing rewards and no visible sign of collapse.
This suggests that multi-rollout sampling can partially mitigate instability, but does not guarantee stable training.
Compared with multi-rollout sampling alone, the grouping mechanism is more critical for training stability -- it ensures each effective group contains both positive and negative samples.
% In other words, even with multiple rollouts, groups containing only positive or only negative samples may still induce unstable training dynamics.

\begin{figure*}[h!]
    \centering
    \includegraphics[width=\linewidth]{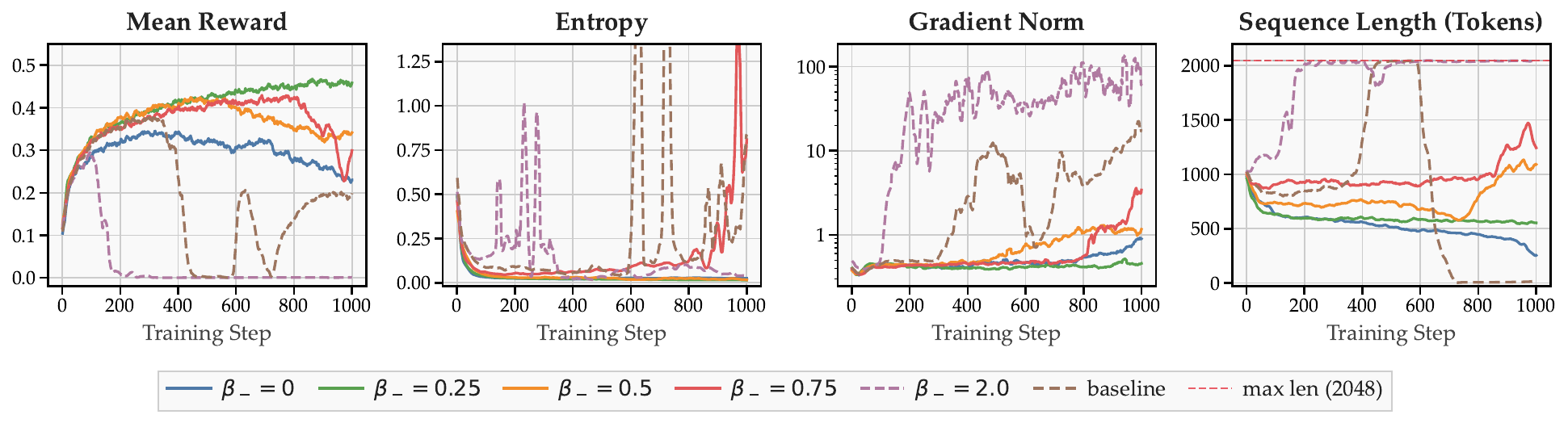}
    \caption{Learning dynamics under varying negative coefficients $\beta_k$. We train the Qwen2.5-Math-1.5B model on a 7.5k-example subset of the DAPO-MATH dataset for 1k steps with a learning rate of $3\times10^{-6}$. Each step contains $512$ prompts and uses a mini-batch size of $32$. We apply an in-reward KL penalty with a coefficient of $1\times10^{-3}$.}
    \label{fig:neg_ablation}
    \vspace{-1em}
\end{figure*}

\subsection{The Impact of the Negative}
\label{sec:negative}

In the previous section, we reveal that the group-based advantage can effectively mitigate training instability.
In this section, we next investigate, in the absence of the grouping,
which component is primarily responsible for instability: positive or negative samples. 

\begin{figure*}
    \centering
    \includegraphics[width=\linewidth]{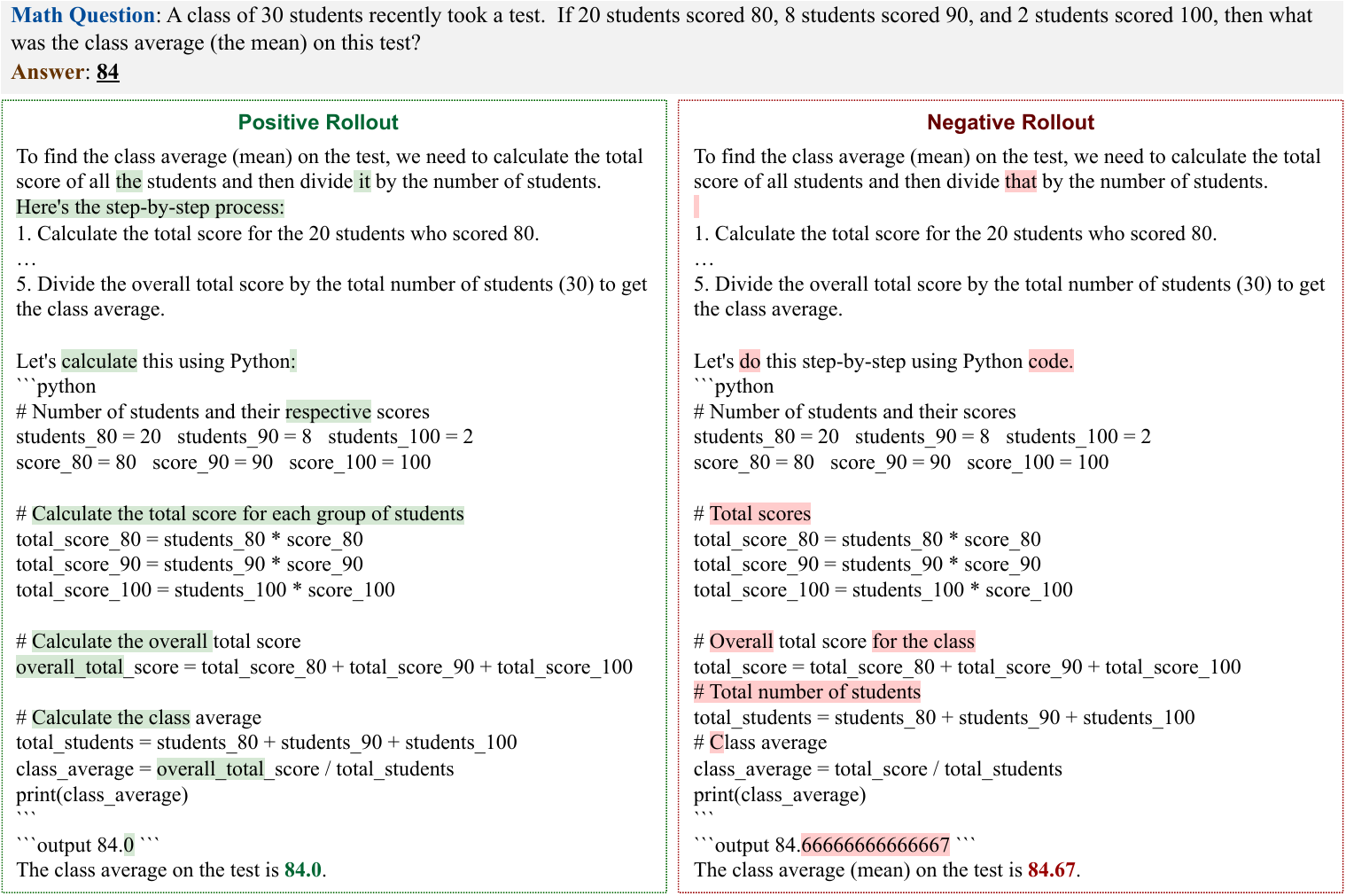}
    \caption{An example illustrating that positive and negative rollouts for the same prompt can share many \emph{\uline{supporting tokens}}, shown as \uline{uncolored text}.}
    \label{fig:supporting_tokens}
\end{figure*}

Trajectories with negative advantages induce updates that decrease the likelihood of the sampled tokens. 
% These updates can be viewed as gradient ascent on negative samples, a mechanism closely related to objectives used in machine unlearning~\cite{zhang2024negative}.
However, especially in RLVR, a negative sample is labeled only by its incorrect final answer. 
As the old saying goes, \emph{Bonum ex integra causa, malum ex quocumque defectu.}~\footnote{Literal translation: \emph{Good arises from an integral cause; evil from any defect whatsoever}.} 
An incorrect final answer does not imply that every token in the trajectory is erroneous or should be penalized. 
We therefore hypothesize that \uline{uniformly penalizing negative trajectories can falsely discourage useful reasoning patterns, grammatical tokens, and other functional components} -- we denote them as \emph{\uline{supporting tokens}} (an example is shown in Fig.~\ref{fig:supporting_tokens}) -- thereby inducing unstable and potentially destructive updates.

\begin{figure}[h!]
    \centering  
    \includegraphics[width=0.95\linewidth]{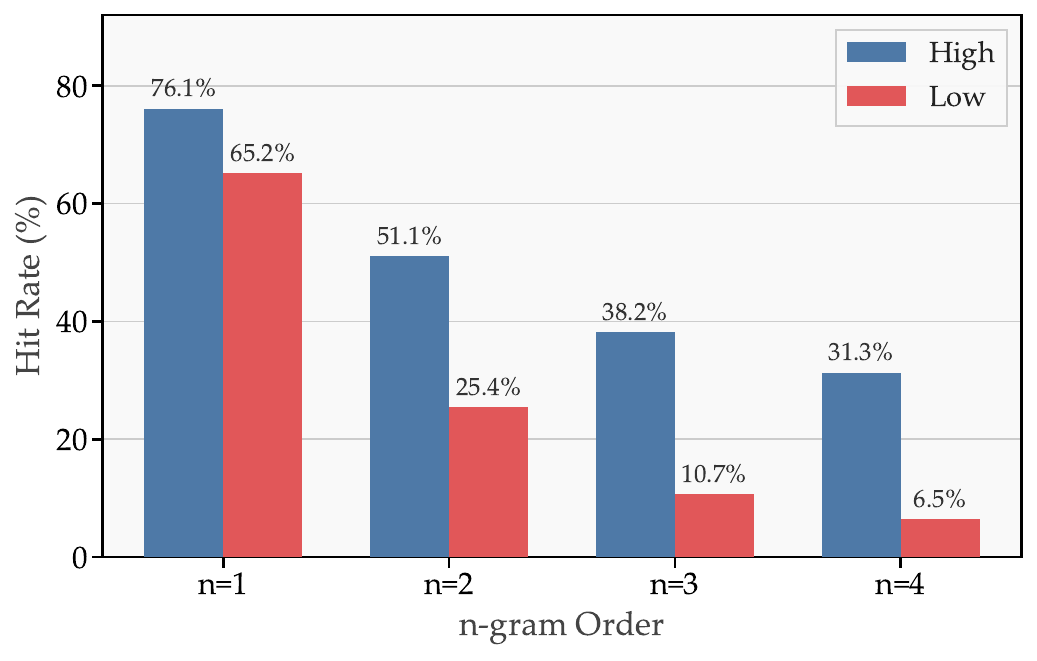}
    \caption{Hit rates in positive rollouts of high- and low-probability $n$-grams from negative rollouts. Results are computed with Qwen2.5-Math-1.5B on MATH500. 
    For each prompt, we generate 10 responses and discard degenerated trajectories. Among 500 groups, 141 groups with all correct or all incorrect responses are excluded.}
    \label{fig:hit_rate}
\end{figure}

To verify our hypothesis, we conduct a controlled experiment on \rfpp\gs{1} that varies the strength of the negative signal while keeping the positive signal fixed. 
Specifically, we scale the negative advantage by a coefficient $\beta_-$ and examine its effect on training dynamics,
monitoring reward, entropy, gradient norm, and sequence length throughout training.
% shown in Fig.~\ref{fig:neg_ablation}. 
% where $\beta_-=1$ denotes the default setting. 
As shown in Fig.~\ref{fig:neg_ablation}, collapse manifests as a sharp drop in reward, accompanied by simultaneous increases in entropy, gradient norm, and sequence length.
Once the sequence length saturates at the preset maximum of 2048 tokens, 
training briefly stabilizes before collapsing again. 
During this transient phase, the KL penalty becomes the dominant term in the objective, driving the sequence length back down. 
By this point, the model has already suffered substantial degradation.

As we decrease $\beta_-$ from $1$ (the baseline) toward $0$, 
training becomes progressively more stable, and the onset of collapse is delayed.~\footnote{Note that collapse is stochastic: even under the same configuration, it may occur at different training steps. However, the overall trend is clear.}  
At $\beta_-=0.25$, training remains stable throughout the full 1k-step training horizon: the reward increases steadily, with no abrupt drops, while entropy, gradient norm, and sequence length remain well-behaved. 
Since this horizon is substantially longer than the typical collapse time observed under larger negative coefficients, the stability is unlikely to be incidental.
The opposite further supports this conclusion: when $\beta_-=2$, corresponding to the strongest negative signal we test, 
collapse occurs earlier than in all other configurations. 
Together, these results indicate that training instability is primarily driven by negative samples rather than positive samples.

\subsection{Supporting Tokens in Rollouts}

Motivated by the previous findings -- the instabilities stem from negative samples, whereas the positive rollouts from the same group can mitigate --,
we hypothesize that:
1) in the absence of positive rollouts, negative samples lead to penalties on supporting tokens, inducing harmful updates;
2) when optimized jointly, positive samples can offset these harmful gradients through their shared supporting tokens~\cite{cheng2026cancellation}.

This hypothesis is empirically verified by measuring token-level overlap between positive and negative samples generated from the same prompt (shown in Fig.~\ref{fig:hit_rate}).
Notably, we further categorize the tokens into two types: 
high-probability tokens as those whose probabilities fall within the top $90\%$ in a sequence, 
and define the remaining tokens as low-probability tokens.
As shown in Fig.~\ref{fig:hit_rate}, high-probability $n$-grams consistently achieve higher hit rates than low-probability $n$-grams across different values of $n$.
This evidence supports two key findings: 
1) positive samples within the same group protect training stability by offsetting harmful updates induced by negative samples on shared supporting tokens; 2) these supporting tokens, particularly at the $n$-gram level, are more likely to fall into the high-probability token category.

% We hypothesize that grouping improves stability because positive samples from the same prompt contain many of these same functional tokens, allowing their negative updates to be partially cancelled during grouped optimization.

% \textcolor{purple}{Liheng: shall we have an visualization for: the cancel-out cannot be done by cross-prompt positive-negative pairs}

% In the one-rollout setting, 
% however, 
% there are no positive samples within the same group to offset the harmful gradients on supporting tokens.
% Following Sec.~\ref{sec:negative}, we therefore attribute this instability to the indiscriminate penalization on the supporting tokens in negative samples.

This observation aligns with a natural intuition: tokens assigned higher probabilities by the policy are more likely to be supporting tokens, as the pretrained base model already encodes strong language-modeling and reasoning priors. 
This insight also suggests a straightforward mitigation for single-rollout RL methods: 
suppress supporting tokens in negative samples with prediction confidence as a proxy,
motivating our proposed technique -- NTF -- discussed in Section~\ref{sec:ntf}.

\subsection{Top-$K$ Subspace Alignment}

In the previous section, we have statistically shown shared supporting tokens among the positive and negative rollouts within a group.
The failure mode is not a gradual decline in reward but an abrupt collapse of the model's language and reasoning ability---the signature of catastrophic forgetting~\cite{kirkpatrick2017overcoming}.
We hypothesize that a layer's pretrained competence is carried mainly by the dominant directions of its weight matrix, i.e., its top singular subspaces, so that an update is harmful to the extent that its gradient concentrates there.
This motivates a direct question: for each type of update, how much of the gradient's energy falls within the top singular subspace of the weights?
We answer it by projecting the gradient onto the top-$k$ singular subspaces, as described below.

For a weight matrix $\bm{W}\in\R^{d_{\text{out}}\times d_{\text{in}}}$, we compute its singular value decomposition:
\begin{equation}
\bm{W} = \bm{U}\bm{\Sigma}\bm{V}^\intercal,
\end{equation}
where $\bm{U}\in\R^{d_{\text{out}}\times r}$ and
$\bm{V}\in\R^{d_{\text{in}}\times r}$ have orthonormal columns,
$\bm{\Sigma}=\operatorname{diag}(\sigma_1,\dots,\sigma_r)$ with
$\sigma_1\ge\cdots\ge\sigma_r\ge 0$, and
$r=\min(d_{\text{out}},d_{\text{in}})$. Let $\bm{U}_k$ and $\bm{V}_k$ contain the leading $k$ left and right singular vectors, respectively; these vectors span the top-$k$ left and right singular subspaces of $\bm{W}$.

Given a gradient $\bm{G}\in\R^{d_{\text{out}}\times d_{\text{in}}}$ with the same shape as $\bm{W}$, we project it onto the top-$k$ singular subspaces of $\bm{W}$:
\begin{equation}
\bm{P}_k = \bm{U}_k^\intercal \bm{G}\bm{V}_k
\;\in\; \R^{k\times k}.
\end{equation}
This projection is the leading $k\times k$ block of $\bm{G}$ expressed in the singular-vector basis of $\bm{W}$. We then define the normalized block energy as
\begin{equation}
\rho_k =
\frac{\lVert \bm{P}_k \rVert_F^2}{\lVert \bm{G} \rVert_F^2}
\;\in\; [0,1],
\end{equation}
which measures the fraction of the gradient's Frobenius energy captured by the top-$k$ singular subspaces. Since orthogonal transformations preserve the Frobenius norm, $\rho_k$ is non-decreasing in $k$ and reaches $1$ at $k=r$. A large $\rho_k$ for small $k$ therefore indicates that the gradient is concentrated along the dominant directions of $\bm{W}$.

\begin{figure}[h!]
    \centering
    \includegraphics[width=0.95\linewidth]{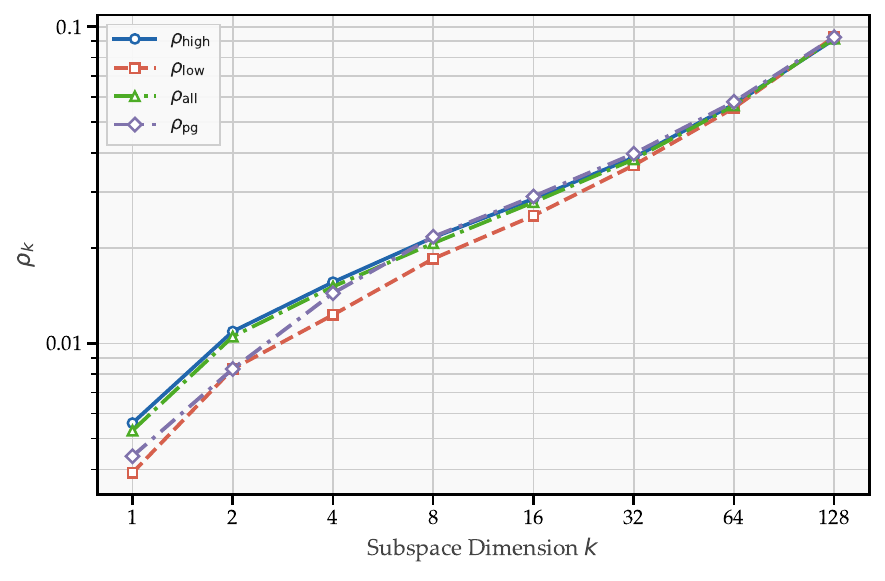}
    \caption{Block energy $\rho_k$ as a function of the subspace dimension $k$,
    computed with Qwen2.5-Math-1.5B on MATH500 over $1{,}380$ positive and
    $1{,}380$ negative samples drawn from $359$ prompts. Each curve is averaged
    over $197$ weight matrices.}
    \label{fig:rho_k}
\end{figure}

We compute the gradient from the advantage-weighted log-likelihood loss
\begin{equation}
    l = -\frac{A}{|o|}\sum_{t=1}^{|o|}
        m_t \,\log \pi_\theta\!\left(o_t \mid o_{<t}, q\right),
    \,\,
    \bm{G} = \frac{\partial l}{\partial \bm{W}},
\end{equation}
where $o$ is the sampled trajectory, $q$ the question, $|o|$ the trajectory length, $A$ the advantage, and $m_t\in\{0,1\}$ a token mask that selects which tokens contribute to the gradient. We compare four settings. The first three isolate the effect of a token subset: we fix $A=1$, accumulate the gradient over both positive and negative samples, and vary only the mask---(1) \textbf{high-prob.}\ tokens, (2) \textbf{low-prob.}\ tokens, and (3) \textbf{all tokens} ($m_t=1$). The fourth, (4) \textbf{policy gradient}, sets $m_t=1$ and uses the group-relative advantage ($A=1$ for correct and $A=-1$ for incorrect trajectories), so that positive and negative trajectories enter with opposite signs.

Fig.~\ref{fig:rho_k} reveals three patterns:
(i) the all-token curve closely tracks the high-prob. token curve;
(ii) for small $k$, high-prob. tokens attain a substantially larger
$\rho_k$ than low-prob. tokens; and
(iii) at $k=1,2$, the policy-gradient curve nearly coincides with the
low-prob. token curve.

These patterns support our hypothesis. High-prob. tokens align far more strongly with the top singular directions than low-prob. tokens (observation~(ii)), so updating on them perturbs precisely the directions we associate with the model's core capability---and since the all-token gradient is dominated by these tokens (observation~(i)), naive single-rollout training cannot avoid this. Under group-based optimization, however, the high-prob. gradients of positive and negative trajectories are top-$K$-aligned but carry opposite-sign advantages, and therefore largely cancel; the policy-gradient curve consequently collapses onto the low-prob. curve at $k=1,2$ (observation~(iii)). Grouping thus shields the dominant directions from over-penalization, preventing collapse.

\section{Methodology}

% Our method has two components: an asymmetric clipped objective that decouples
% positive and negative trajectories (Section~\ref{sec:objective}), and a negative
% token-filtering scheme that prevents unstable updates on high-probability tokens
% within negative trajectories (Section~\ref{sec:ntf}).

\subsection{Batch-level Advantage Computation}
\label{sec:objective}

\paragraph{Notation.}
Let $\pi_\theta$ be the current policy with parameter $\theta$ and
$\pi_{\theta_\text{old}}$ the behavior policy used to collect rollouts, and let
$\rho(o_t) = \pi_\theta(o_t)/\pi_{\theta_\text{old}}(o_t)$ denote the per-token
importance ratio. For a query $q$, each sampled trajectory is labeled positive
($o^+$) or negative ($o^-$); in RLVR with binary verifiable rewards this label is
simply the correctness of the trajectory. We write $N^+$ and $N^-$ for the number
of positive and negative trajectories in a mini-batch.
For continuous rewards, the positive and negative samples are determined by the value baseline estimated by group/batch average.

\paragraph{Base RL techniques with batch-level advantage estimation.}
We mainly verify our proposed techniques on two RL techniques with batch-level advantage estimation.

\textbf{REINFORCE++ (RF++).} RF++~\cite{hu2025reinforce++} forms the advantage by z-score normalizing rewards within a generation batch $B$,
\begin{equation}
    A = \frac{r - \mu_B}{\mathrm{std}_B}.
\end{equation}

\textbf{Contrastive-REINFORCE (C-RF).}
For binary rewards with batch success rate $p$, 
we have $\mu_B = p$ and
$\mathrm{std}_B = \sqrt{p(1-p)}$. \citet{wu2025takes}
and~\citet{li2026disco} show that the surrogate with z-score normalization yields~\footnote{Conditioning on the outcome, $A^+=\sqrt{(1-p)/p}$ with
probability $p$ and $A^-=-\sqrt{p/(1-p)}$ with probability $1-p$; multiplying by
the respective outcome probabilities gives the shared coefficient
$p\,A^+ = -(1-p)\,A^- = \sqrt{p(1-p)}$.}
\begin{equation}
\label{eq:z_score_obj}
    \mathcal{J} = \sqrt{p(1-p)}\;
    \mathbb{E}_{q}\!\left[\, l^+(o^+\!\mid q) - l^-(o^-\!\mid q) \,\right],
\end{equation}
where the scalar $\sqrt{p(1-p)} = \mathrm{std}_B$ is the \emph{standard deviation}
of the batch success indicator,
which might lead to an undesired weighting effect in RLVR setting -- 
amplifying at $p=\tfrac{1}{2}$
and attenuating  as $p\to 0$ or $p\to 1$.
This might be more pronounced in agentic settings, where the initial success rate is near zero.

Therefore, we also verify on another objective, W-REINFORCE~\cite{zhu2026surprising}, without the weighting coefficient:
\begin{equation}
\begin{aligned}
    &\mathcal{L} = -\frac{1}{2}\left(
        \frac{1}{N^+}\sum_{i=1}^{N^+} l^+(o_i^+)
        \;-\;
        \frac{1}{N^-}\sum_{j=1}^{N^-} l^-(o_j^-)
    \right), \\
    &l^+(o) = \frac{1}{|o|}\sum_{t=1}^{|o|}
        \min\!\big(\rho(o_t),\, 1+\epsilon\big), \\
    &l^-(o) = \frac{1}{|o|}\sum_{t=1}^{|o|}
        \max\!\big(\rho(o_t),\, 1-\epsilon\big).
\end{aligned}
\end{equation}
We denote it as Contrastive-REINFORCE (C-RF), as the weighting coefficient is removed.

\subsection{Negative Token Filtering (NTF)}
\label{sec:ntf}
A negative trajectory is penalized token by token, but not every token is to
blame for the failure. Many of its high-probability tokens are generic or are
also produced during correct behavior.
Penalizing them
indiscriminately injects noise into the update and destabilizes training. We
therefore concentrate the negative gradient on the tokens most plausibly
responsible for the failure.

Concretely, given a masking fraction $\tau\in[0,1]$, we rank the tokens of each
negative trajectory by their probability under the current policy and mask the
top-$\tau$ fraction (the highest-probability tokens). Let $\mathcal{K}(o)$ denote
the set of tokens that remain. The negative term is still normalized by the full
sequence length,
\begin{equation}
    l^-(o) = \frac{1}{|o|}
    \sum_{t \in \mathcal{K}(o)} \max\!\big(\rho(o_t),\, 1-\epsilon\big),
\end{equation}
so masked tokens contribute nothing while still counting toward the denominator.
This shields high-probability tokens from spurious penalties while keeping the
update focused on the tokens that drive the negative outcome.
\section{Related Work}

\paragraph{Group-Based RL Algorithms}
Most current RL algorithms for LLMs are group-based; they require multiple rollouts per prompt.
GRPO and RLOO are two representative members of this family. ReMax uses two rollouts per prompt and can therefore also be regarded as group-based.
We note that REINFORCE++ performs a single rollout in the RLHF setting but multiple rollouts in the RLVR setting.
In all of these methods, the group of rollouts is used to estimate the value function, thereby reducing variance. Recent work offers an alternative interpretation in which group-based methods are viewed as a form of contrastive learning~\cite{wu2025takes, zhu2026surprising}, where the primary role of grouping is to supply contrastive pairs.

\paragraph{Group-Free RL Algorithms}
PPO~\cite{schulman2017proximal}, the most widely used RL algorithm, requires only a single rollout per prompt. However, it must maintain a separate LLM as the critic. More recently, SPO has been proposed as a critic-free, group-free method. SPO leverages historical information to estimate the success rate of each prompt, which serves as the value function. To obtain accurate estimates, SPO requires a pre-rollout phase before training, incurring additional computational cost. In this paper, we instead focus on uncovering the hidden mechanism behind grouping. Our proposed negative filtering enables stable single-rollout training without any additional computational overhead.

\paragraph{Token Masking in RL}

\citet{wang2025beyond} propose improving RL post-training by only optimizing high-entropy minority tokens. 
While this idea is philosophically related to our proposed NTF, the two methods differ in both motivation and implementation. 
NTF applies only to negative trajectories, whereas \citet{wang2025beyond} are not restricted to negative samples and targets high-entropy tokens.
\section{Experiment}
\label{sec:experiment}

In this section, we conduct empirical experiments to validate the efficacy of our proposed NTF on two RL base techniques with batch-level advantage estimation: RF++ and C-RF.

\paragraph{Experimental Setup}
We evaluate on mathematical reasoning and agentic decision-making. The reasoning experiments train on a 7.5K-prompt subset of DAPO-Math \citep{yu2026dapo, wu2025takes} and evaluate on five held-out math benchmarks; the agentic experiments follow \citet{he2026hierarchy} on ALFWorld and WebShop.
We use verl~\cite{sheng2024hybridflow} for experiment framework.
Full baseline, dataset, metric, reward, and hyper-parameter details are provided in Appendix~\ref{app:exp_details}.

\subsection{Reasoning Task}
\label{sec:reasoning_task}

\begin{table*}[t]
\centering
\caption{Mathematical reasoning results under different rollout budgets. All methods are
trained on the 7.5K-prompt DAPO-Math subset and evaluated with Mean@32 accuracy (\%) on five
held-out benchmarks. All $G{=}1$ rows use the same log-probability-based negative-token
filtering ratio $\tau{=}0.1$. ``--'' indicates a training crash.}
\label{tab:rfpp_rollout_compare}
\small
\setlength{\tabcolsep}{5pt}
\renewcommand{\arraystretch}{1.15}
\begin{tabular}{llc ccccc c}
\toprule
& & & \multicolumn{5}{c}{Benchmarks (Mean@32 $\uparrow$)} & \\
\cmidrule(lr){4-8}
Model & Algorithm & $G$ & MATH-500 & AMC 2023 & Minerva & AIME 2025 & Olympiad & Avg. \\
\midrule
\multirow{7}{*}{Qwen2.5-Math-1.5B}
& GRPO        & 16 & 71.04 & 57.11 & 18.22 & 10.00 & 24.86 & 36.25 \\
& GRPO        & 2  & 68.81 & 52.19 & 16.79 & 8.13  & 23.52 & 33.89 \\
\cmidrule(lr){2-9}
& RF++        & 8  & 67.48 & 54.77 & 17.36 & 6.46  & 22.87 & 33.79 \\
& RF++        & 1  & --    & --    & --    & --    & --    & --    \\
\rowcolor{gray!12}
& RF++ w/ NTF & 1  & 68.21 & 55.47 & 16.31 & 2.60  & 21.68 & 32.86 \\
& C-RF        & 1  & --    & --    & --    & --    & --    & --    \\
\rowcolor{gray!12}
& C-RF w/ NTF & 1  & 69.63 & 55.47 & 18.35 & 9.38  & 23.89 & 35.34 \\
\midrule
\multirow{7}{*}{Qwen2.5-Math-7B}
& GRPO        & 16 & 76.94 & 68.91 & 25.67 & 15.94 & 28.66 & 43.22 \\
& GRPO        & 2  & 77.43 & 64.84 & 21.95 & 14.58 & 29.86 & 41.73 \\
\cmidrule(lr){2-9}
& RF++        & 8  & 71.53 & 58.36 & 23.39 & 6.88  & 24.74 & 36.98 \\
& RF++        & 1  & --    & --    & --    & --    & --    & --    \\
\rowcolor{gray!12}
& RF++ w/ NTF & 1  & 64.94 & 68.91 & 21.97 & 6.15  & 24.92 & 37.37 \\
& C-RF        & 1  & --    & --    & --    & --    & --    & --    \\
\rowcolor{gray!12}
& C-RF w/ NTF & 1  & 74.08 & 61.95 & 23.82 & 12.19 & 26.54 & 39.72 \\
\bottomrule
\end{tabular}
\end{table*}

Table~\ref{tab:rfpp_rollout_compare} reports mathematical reasoning results under different rollout budgets. The main observation is that single-rollout training becomes viable when negative trajectories are filtered. 
In our experiments, RF++ and C-RF with $G\!=\!1$ can be trained stably with our proposed NTF, while the same setting without filtering collapses during training. 
This suggests that directly penalizing all tokens in failed trajectories is harmful for long-form reasoning, 
Filtering mitigates this issue by selectively reducing negative updates on high-confidence tokens and thus preserves useful reasoning patterns.

Compared to RF++ w/ NTF,
C-RF w/ NTF reaches a better overall performance,
outperforming the low-budget GRPO baseline with $G=2$ on the 1.5B model and approaches GRPO with $G=16$, with only a 0.91 gap in average for the 1.5B model. 
On the 7B model, group-based GRPO remains stronger, 
but our method still substantially stablize the training of RF++/C-RF,
while using only one rollout per prompt.

We attribute this advantage to a stronger and more reliable training signal in the single-rollout setting: after filtering harmful negative updates on supporting tokens, our objective can make better use of the remaining sparse reward signal than vanilla batch-level advantage computation.

\subsection{Agentic Task}
\label{sec:agentic_task}

\begin{table*}[t!]
\centering
\caption{Agentic decision-making results on ALFWorld and WebShop. We compare prompting
baselines and RL-trained methods on Qwen2.5-1.5B-Instruct and Qwen2.5-7B-Instruct. ALFWorld
reports in-distribution and out-of-distribution success rates, while WebShop reports task
score and task success. ``--'' indicates a training crash.
Baseline results are from~\citet{he2026hierarchy}.}
\label{tab:agentic}
\small
\setlength{\tabcolsep}{6pt}
\renewcommand{\arraystretch}{1.15}
\begin{tabular}{cll cc cc}
\toprule
\multirow{2}{*}{Model} & \multirow{2}{*}{Type} & \multirow{2}{*}{Method}
& \multicolumn{2}{c}{ALFWorld} & \multicolumn{2}{c}{WebShop} \\
\cmidrule(lr){4-5}\cmidrule(lr){6-7}
& & & In-Success & Out-Success & Task Score & Task Success \\
\midrule
\multirow{8}{*}{\rotatebox[origin=c]{90}{\scriptsize Qwen2.5-1.5B-Instruct}}
& Prompting   & Qwen2.5           & 4.1  & --   & 23.1 & 5.2  \\
& Prompting   & ReAct             & 12.8 & --   & 40.1 & 11.3 \\
& Prompting   & Reflexion         & 21.8 & --   & 55.8 & 21.9 \\
& RL Training & PPO (with critic) & 54.4 & --   & 73.8 & 51.5 \\
& RL Training & RLOO              & 69.7 & 68.7 & 73.9 & 52.1 \\
& RL Training & GRPO              & 72.8 & 70.1 & \textbf{75.8} & 56.8 \\
\cmidrule(lr){2-7}
& RL Training & C-RF              & --   & --   & --   & --   \\
\rowcolor{gray!12}
& RL Training & C-RF w/ NTF       & \textbf{87.4} & \textbf{77.9} & 72.4 & \textbf{65.7} \\
\midrule
\multirow{8}{*}{\rotatebox[origin=c]{90}{\scriptsize Qwen2.5-7B-Instruct}}
& Prompting   & Qwen2.5           & 14.8 & --   & 26.4 & 7.8  \\
& Prompting   & ReAct             & 31.2 & --   & 46.2 & 19.5 \\
& Prompting   & Reflexion         & 42.7 & --   & 58.1 & 28.8 \\
& RL Training & PPO (with critic) & 77.1 & 76.2 & \textbf{81.4} & 68.7 \\
& RL Training & RLOO              & 77.9 & 74.0 & 80.3 & 65.7 \\
& RL Training & GRPO              & 78.6 & 76.8 & 79.3 & 66.1 \\
\cmidrule(lr){2-7}
& RL Training & C-RF              & --   & --   & --   & --   \\
\rowcolor{gray!12}
& RL Training & C-RF w/ NTF       & \textbf{90.5} & \textbf{86.3} & 80.2 & \textbf{74.1} \\
\bottomrule
\end{tabular}
\end{table*}

In agentic tasks, we evaluate C-RF with NTF only, as we were unable to successfully train RF++ with NTF in these settings.
We attribute this to the coefficient $\sqrt{p(1-p)}$ discussed in Eq.~\ref{eq:z_score_obj}: because agentic tasks are substantially harder in the early stages of training, $p$ is small, so this coefficient sharply attenuates the learning signal and makes training difficult.

Table~\ref{tab:agentic} reports results on ALFWorld and WebShop, which test whether our single-rollout training principle generalizes beyond math reasoning to agentic tasks under sparse, delayed rewards. RL training substantially outperforms prompting-based methods in both environments, confirming that environment feedback supplies useful supervision for agentic behavior. We compare primarily against PPO, RLOO, and GRPO, representing critic-based and group-based critic-free training.

On ALFWorld, our method is strong at both model scales. Qwen2.5-1.5B-Instruct reaches 87.38 in-distribution and 77.86 out-of-distribution success, exceeding GRPO on both. Qwen2.5-7B-Instruct further improves these to 90.48 and 86.32, surpassing PPO, RLOO, and GRPO.
% The single-rollout objective is thus effective in long-horizon settings where reward is observed only after a complete trajectory.

WebShop shows a similar advantage on task success: our method raises success over GRPO from 56.8 to 65.73 (1.5B) and from 66.1 to 74.07 (7B). Results on task score are more mixed—our method trails GRPO on the 1.5B model and PPO/RLOO on the 7B model, though it remains competitive.

The stronger performance of C-RF w/ NTF on agentic tasks suggests that group-based cancellation may be less reliable in complex search spaces, 
where positive and negative trajectories share relatively fewer supporting tokens due to the large search spaces.
This further highlights the importance of selectively controlling negative-sample updates. 
Extending NTF to group-based RL algorithms is a promising direction for future work.

\subsection{Sensitivity Study}
\begin{figure*}[ht!]
    \centering
    \includegraphics[width=\linewidth]{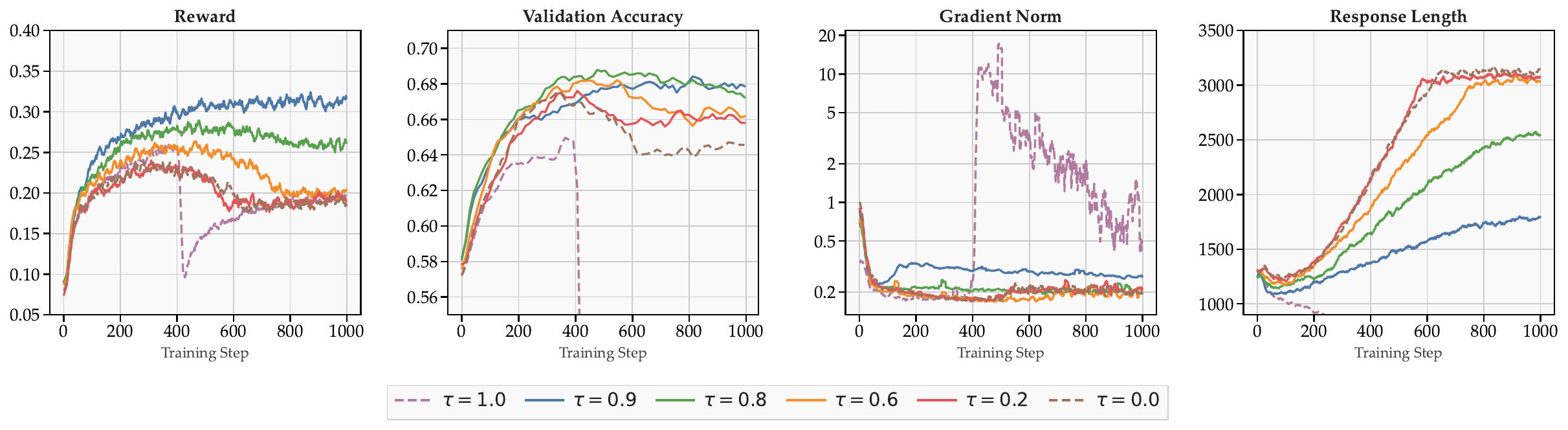}
    \caption{Learning dynamics under varying negative masking fractions $\tau$. In each setting, Qwen2.5-Math-1.5B is trained with C-RF for 1k steps.}
    \label{fig:sensity_study}
\end{figure*}

To study the effect of the masking fraction $\tau$ on training stability, we conduct a sensitivity analysis of this key hyperparameter in negative token filtering, which determines the proportion of negative tokens retained.
In the study, we follow the the setting of the reasoning tasks -- the base model, Qwen2.5-Math-1.5B, is trained on a DAPO subset with C-RF for 1k steps. 
Compared with the previous experiment (Fig.~\ref{fig:neg_ablation}), we adopt a more benign learning rate ($1\times10^{-6}$) with a cosine scheduler and no KL penalty. As shown in Fig.~\ref{fig:sensity_study}, we report the curves of reward, validation accuracy, gradient norm, and response length over training; we omit entropy, as it provides little additional information.
We observe that: (1) under this benign learning-rate setting, collapse manifests as a degeneration in reward and validation accuracy; and (2) response length is strongly correlated with this degeneration. For $\tau \leq 0.6$, the degeneration is pronounced, whereas for $\tau = 0.9$ and $0.8$ training remains stable. Moreover, $\tau = 0.9$, our default choice, achieves higher reward and shorter responses than $\tau = 0.8$.
However, we note that a larger $\tau$ does not always yield better performance. 
% To verify this, we additionally conduct an experiment with 
In the setting with $\tau = 1$ (positive-only), 
a different type of collapse occurs: the response length drops rapidly, and the model forgets to reason, generating the answer directly instead.
These results demonstrate the importance and necessity of appropriate negative token filtering.
\section{Conclusion}

We revisited the role of multi-rollout sampling and group-based advantage in critic-free RLVR and showed that their function extends beyond value-baseline estimation. 
In particular, groups stabilize training by allowing positive and negative rollouts from the same prompt to cancel harmful negative updates on shared supporting tokens.
This perspective explains why single-rollout training is especially vulnerable to negative samples, which can falsely penalize useful reasoning patterns, grammatical tokens, and structural components.

Motivated by this insight, we proposed Negative Token Filtering (NTF), which masks high-probability tokens in negative trajectories and concentrates the negative update on lower-probability tokens. 
NTF enables stable single-rollout, critic-free, and group-free training. 
Experiments on mathematical reasoning and agentic decision-making show that our method avoids collapse and achieves competitive or stronger performance than representative group-based baselines. 
These results suggest that stable RLVR does not inherently require group-based advantage computation, provided that negative-token updates are handled selectively.


\clearpage
\newpage
\section{Limitation}
Our analysis is primarily static: we do not conduct dynamic experiments that track metrics throughout RL training. This is because our weight analysis relies on SVD, which is computationally expensive even at small scale, making spectrum analysis over the course of training impractical. In addition, our experiments focus on the RLVR (binary reward) setting and do not cover ordinal or continuous rewards. We believe RLVR is representative enough to validate our findings, and we leave the exploration of continuous-reward settings to future work.

We use LLMs for polishing writing.
\bibliography{ref}

\clearpage
\newpage
\appendix

\section{Experimental Details}
\label{app:exp_details}

\subsection{Datasets}
\label{app:dataset_details}
For mathematical reasoning, all RL methods are trained on the same 7.5K-prompt subset of DAPO-Math~\citep{yu2026dapo} and evaluated on MATH-500~\cite{hendrycks2021measuring, lightman2023letsverifystepstep}, AMC 2023~\cite{amc23_hf}, Minerva Math~\cite{lewkowycz2022solvingquantitativereasoningproblems}, AIME 2025~\cite{matharena_aime2025}, and Olympiad Bench~\cite{he2024olympiadbenchchallengingbenchmarkpromoting}. We use binary verifiable rewards based on extracted final-answer correctness and report Mean@32 accuracy, computed from 32 sampled generations per problem.

For agentic tasks, we follow the ALFWorld~\cite{shridhar2021alfworldaligningtextembodied} and WebShop~\cite{yao2023webshopscalablerealworldweb} protocol of \citet{he2026hierarchy}. We train our method on a 1024-task ALFWorld subset and a 1024-instruction WebShop subset, and evaluate with the same metrics as the reported baselines: in-/out-of-distribution success for ALFWorld, and task score/task success for WebShop. Rewards are sparse and delayed environment feedback with no per-step shaping; for WebShop, we use binary task success as the training reward.

\subsection{Baseline Details}
\label{app:baseline_details}

We provide additional details about the baselines used in our experiments. The baselines are selected to compare our method with representative RLVR~\cite{lambert2025tulu3pushingfrontiers} training paradigms under different rollout budgets. In particular, we use REINFORCE++~\cite{hu2025reinforce++} with $G=1$ as the closest single-rollout baseline, GRPO~\cite{shao2024deepseekmath} with small and large group sizes as group-based references, and standard prompting/RL baselines for agentic tasks.

\paragraph{Mathematical reasoning.}
For mathematical reasoning, we compare with GRPO and REINFORCE++. GRPO is a representative group-based critic-free RLVR method, where multiple responses are sampled for the same prompt and relative advantages are estimated within each group. We report GRPO with $G=16$ as a strong multi-rollout baseline and GRPO with $G=2$ as a lower-budget group-based baseline. These two settings allow us to examine how our single-rollout method compares with group-based training under different rollout costs.

REINFORCE++ is used as the main batch-normalized critic-free baseline. We report REINFORCE++ with $G=8$ as a multi-rollout batch-normalized baseline and REINFORCE++ with $G=1$ as the closest comparison to our method. The $G=1$ setting is especially important because it shares the same one-rollout-per-prompt constraint as our method. For all single-rollout runs conducted by us, including REINFORCE++ with $G=1$ and our method, we apply the same log-probability-based negative-token filtering with ratio $\tau=0.1$. Therefore, the comparison between REINFORCE++ with $G=1$ and our method evaluates whether our single-rollout training rule can make better use of the same filtered reward signal, rather than merely testing the effect of adding token filtering.

\paragraph{Agentic tasks.}
For agentic tasks, we compare with both prompting-based and RL-based baselines. The prompting baselines include direct Qwen2.5~\cite{qwen2025qwen25technicalreport} prompting, ReAct~\cite{yao2023reactsynergizingreasoningacting}, and Reflexion~\cite{shinn2023reflexionlanguageagentsverbal}, which measure the performance of the base model without RL training. The RL baselines include PPO~\cite{schulman2017proximal}, RLOO~\cite{ahmadian2024back}, and GRPO. PPO represents critic-based RL training, while RLOO and GRPO represent critic-free methods based on relative or grouped trajectory comparison.

The baseline numbers for agentic tasks are taken from \citet{he2026hierarchy}, which reports results on ALFWorld and WebShop using Qwen2.5-Instruct backbones. Specifically, ALFWorld is evaluated by in-distribution and out-of-distribution success rates, while WebShop is evaluated by task score and task success rate. Our method is evaluated under the same task protocol and metrics for comparison. Unlike the group-based baselines, our method uses only one rollout per prompt and does not require a critic, a value tracker, or prompt-level rollout groups.

\subsection{Hyper-parameters}
\label{app:hyperparam_details}
See Table~\ref{tab:train_hparams} and Table~\ref{tab:eval_hparams}.

\begin{table*}[!t]
  \centering
  \caption{Training hyper-parameters of our method. Values that are shared
           across all three tasks are merged into a single cell; columns
           differ only where shown.}
  \label{tab:train_hparams}
  \small
  \setlength{\tabcolsep}{6pt}
  \renewcommand{\arraystretch}{1.15}
  \begin{tabular}{@{}lccc@{}}
  \toprule
  \textbf{Hyper-parameter} & \textbf{Math} & \textbf{ALFWorld} & \textbf{WebShop} \\
  \midrule
  \multicolumn{4}{@{}l}{\textit{Backbone}} \\
  Base model & Qwen2.5-Math-\{1.5,\,7\}B & \multicolumn{2}{c}{Qwen2.5-\{1.5,\,7\}B-Instruct} \\
  \addlinespace[3pt]
  \multicolumn{4}{@{}l}{\textit{Optimizer}} \\
  Optimizer         & \multicolumn{3}{c}{AdamW} \\
  Learning rate     & \multicolumn{3}{c}{$10^{-6}$} \\
  LR scheduler      & Cosine & Exponential & Exponential \\
  Warm-up steps     & \multicolumn{3}{c}{10} \\
  Weight decay      & \multicolumn{3}{c}{0.1} \\
  Gradient clipping & \multicolumn{3}{c}{1.0} \\
  \addlinespace[3pt]
  \multicolumn{4}{@{}l}{\textit{RL objective}} \\
  Advantage estimator                & RF++, C-RF & C-RF & C-RF \\
  Rollouts per prompt ($G$)          & \multicolumn{3}{c}{1} \\
  Clip $\epsilon_{\text{low}}$       & \multicolumn{3}{c}{0.2} \\
  Clip $\epsilon_{\text{high}}$      & \multicolumn{3}{c}{10} \\
  Negative-token filter ratio $\tau$ & \multicolumn{3}{c}{0.1} \\
  KL penalty in reward               & \multicolumn{3}{c}{disabled} \\
  KL loss in objective               & \multicolumn{3}{c}{disabled} \\
  \addlinespace[3pt]
  \multicolumn{4}{@{}l}{\textit{Batching \& sequence lengths}} \\
  Rollout batch size (prompts) & 512  & 64               & 64 \\
  PPO mini-batch size          & 128  & 64               & 64 \\
  Max prompt length            & \multicolumn{3}{c}{2048} \\
  Max response length          & 4096 & 4096 (all turns) & 4096 (all turns) \\
  \addlinespace[3pt]
  \multicolumn{4}{@{}l}{\textit{Rollout sampling}} \\
  Temperature & \multicolumn{3}{c}{1.0} \\
  Top-$p$     & \multicolumn{3}{c}{1.0} \\
  \addlinespace[3pt]
  \multicolumn{4}{@{}l}{\textit{Environment / reward}} \\
  Reward type           & verifiable binary & binary success & binary success \\
  Max steps per episode & --                & 50             & 15 \\
  \bottomrule
  \end{tabular}
\end{table*}

\begin{table*}[!t]
  \centering
  \caption{Evaluation hyper-parameters.}
  \label{tab:eval_hparams}
  \small
  \setlength{\tabcolsep}{6pt}
  \renewcommand{\arraystretch}{1.2}
  \begin{tabularx}{\textwidth}{@{}l *{3}{>{\raggedright\arraybackslash}X}@{}}
  \toprule
  \textbf{Hyper-parameter} & \textbf{Math} & \textbf{ALFWorld} & \textbf{WebShop} \\
  \midrule
  Sampling temperature & 0.7 & 1.0 & 1.0 \\
  Top-$p$              & 0.7 & 0.7 & 0.7 \\
  Samples per prompt   & 32  & 3   & 3 \\
  Reported metric      & Mean@32 accuracy & In/out-distribution success rate & Task score, success rate \\
  Eval set             & MATH-500, AMC 23, Minerva, AIME 25, Olympiad & ALFWorld in/out-distribution splits & WebShop test set \\
  \bottomrule
  \end{tabularx}
\end{table*}

\end{document}